\documentclass{article}

\usepackage{arxiv}

\usepackage[utf8]{inputenc} 
\usepackage[T1]{fontenc}    
\usepackage{hyperref}       
\usepackage{url}            
\usepackage{booktabs}       
\usepackage{amsfonts}       
\usepackage{nicefrac}       
\usepackage{microtype}      
\usepackage{lipsum}
\usepackage{graphicx}
\usepackage{natbib}
\usepackage{amsmath}
\graphicspath{ {./figures/} }

\title{Target noise: A pre-training based neural network initialization for efficient high resolution learning}

\author{
 Shaowen Wang \\
  King Abdullah University of Science and Technology \\
  Thuwal, Saudi Arabia \\
  \texttt{shaowen.wang@kaust.edu.sa} \\
   \And
 Tariq Alkhalifah \\
  King Abdullah University of Science and Technology \\
  Thuwal, Saudi Arabia \\
  \texttt{tariq.alkhalifah@kaust.edu.sa} \\
}

\begin{document}
\maketitle
\begin{abstract}
Weight initialization plays a crucial role in the optimization behavior and convergence efficiency of neural networks.
Most existing initialization methods, such as Xavier and Kaiming initializations, rely on random sampling and do not exploit information from the optimization process itself.
We propose a simple, yet effective, initialization strategy based on self-supervised pre-training using random noise as the target.
Instead of directly training the network from random weights, we first pre-train it to fit random noise, which leads to a structured and non-random parameter configuration.
We show that this noise-driven pre-training significantly improves convergence speed in subsequent tasks, without requiring additional data or changes to the network architecture.
The proposed method is particularly effective for implicit neural representations (INRs) and Deep Image Prior (DIP)-style networks, which are known to exhibit a strong low-frequency bias during optimization.
After noise-based pre-training, the network is able to capture high-frequency components much earlier in training, leading to faster and more stable convergence.
Although random noise contains no semantic information, it serves as an effective self-supervised signal (considering its white spectrum nature) for shaping the initialization of neural networks.
Overall, this work demonstrates that noise-based pre-training offers a lightweight and general alternative to traditional random initialization, enabling more efficient optimization of deep neural networks.
\end{abstract}


\section{Introduction}
The optimization behavior of deep neural networks is highly sensitive to parameter initialization.
A well-chosen initialization can substantially accelerate convergence, stabilize early-stage optimization, and improve robustness to hyperparameter choices, whereas an unfavorable initialization may lead to slow convergence or even optimization failure.
Motivated by this observation, a lot of research have been devoted to the design of initialization schemes, including Xavier \cite{glorot2010understanding} and Kaiming \cite{He2015Delving} initializations, among many other.
These methods aim to preserve signal variance across layers and alleviate vanishing or exploding gradients, and have proven effective in many practical settings.
Nevertheless, they are fundamentally stochastic and do not explicitly exploit any structural information induced by the network architecture or the optimization process itself prior to training.

At the same time, a growing number of works have shown that neural networks can acquire nontrivial internal structures even in the absence of task-specific targets \cite{doersch2015unsupervised,chen2020simple}.
These studies suggest that the learning process itself can shape the geometry of the parameter space before a target task is introduced.
However, most existing research efforts primarily evaluate such learning strategies based on final task performance, while their impact on optimization dynamics remains largely overlooked.
In particular, how pre-training procedures influence convergence speed, early-stage training behavior, and spectral bias during subsequent optimization has received limited attention.
This leaves open an important question: whether simple, task-agnostic training objectives can be exploited as an effective mechanism for constructing favorable initializations that improve optimization efficiency.

In this paper, we revisit the role of randomness in neural network initialization from the perspective of optimization.
Instead of viewing random noise merely as an unstructured source for stochastic initialization, we examine whether noise itself can be used as a simple training signal to shape the initial parameter configuration of neural networks before solving a target task.
By pre-training the network to fit random noise, the resulting weights depart from purely random states and encode structural properties induced by the network architecture and optimization process.

Related ideas have been explored in \citep{cheon2025pretraining}, where noisy paired data are used as input and target during pre-training, together with a biologically motivated learning rule based on feedback alignment \cite{lillicrap2016random}.
In contrast, our work focuses on standard backpropagation and studies noise-driven pre-training as a general and lightweight initialization strategy.
We systematically analyze its effects on optimization dynamics across different network architectures and single-instance tasks, with particular focus on convergence behavior and spectral properties during training.

Despite the absence of any semantic structure in the noise targets, we find that noise-based pretraining consistently leads to substantially faster convergence than conventional random initialization.
Across a wide range of network architectures and learning tasks, networks initialized with noise-pretrained weights exhibit faster loss decay, more stable early-stage optimization, and reduced low-frequency bias.
Importantly, these improvements are obtained without using additional data, modifying the network architecture, or designing task-specific pretraining objectives, underscoring the simplicity and general applicability of the proposed initialization strategy.

To better understand this phenomenon, we analyze how noise-driven pretraining influences the optimization dynamics of neural networks.
Our results suggest that training on random noise reshapes the parameter initialization into a structured state, which conditions gradient propagation and alters the functional geometry encountered during subsequent optimization.
From a theoretical perspective, this behavior is consistent with recent analyzes of over-parameterized networks, which show that the choice of initialization can strongly affect convergence rates and training trajectories through its impact on the neural tangent kernel (NTK) and its spectral properties \cite{jacot2018neural,du2019gradient}.

Our contributions can be summarized as follows:
\begin{enumerate}
    \item We propose a simple noise-target pretraining strategy for neural network initialization that requires no additional data, architectural modifications, or task-specific objectives.
	
    \item Through several experiments, we show that networks initialized with noise-pretrained weights consistently exhibit faster convergence and more stable early-stage optimization than those using standard random initializations.
	
    \item We analyze the effects of noise-based pretraining on optimization dynamics and provide mechanistic insights into how it reshapes the training process through structured initialization.
    
\end{enumerate}

\begin{figure*}[htbp]
    \centering
    \includegraphics[width=1\textwidth]
    {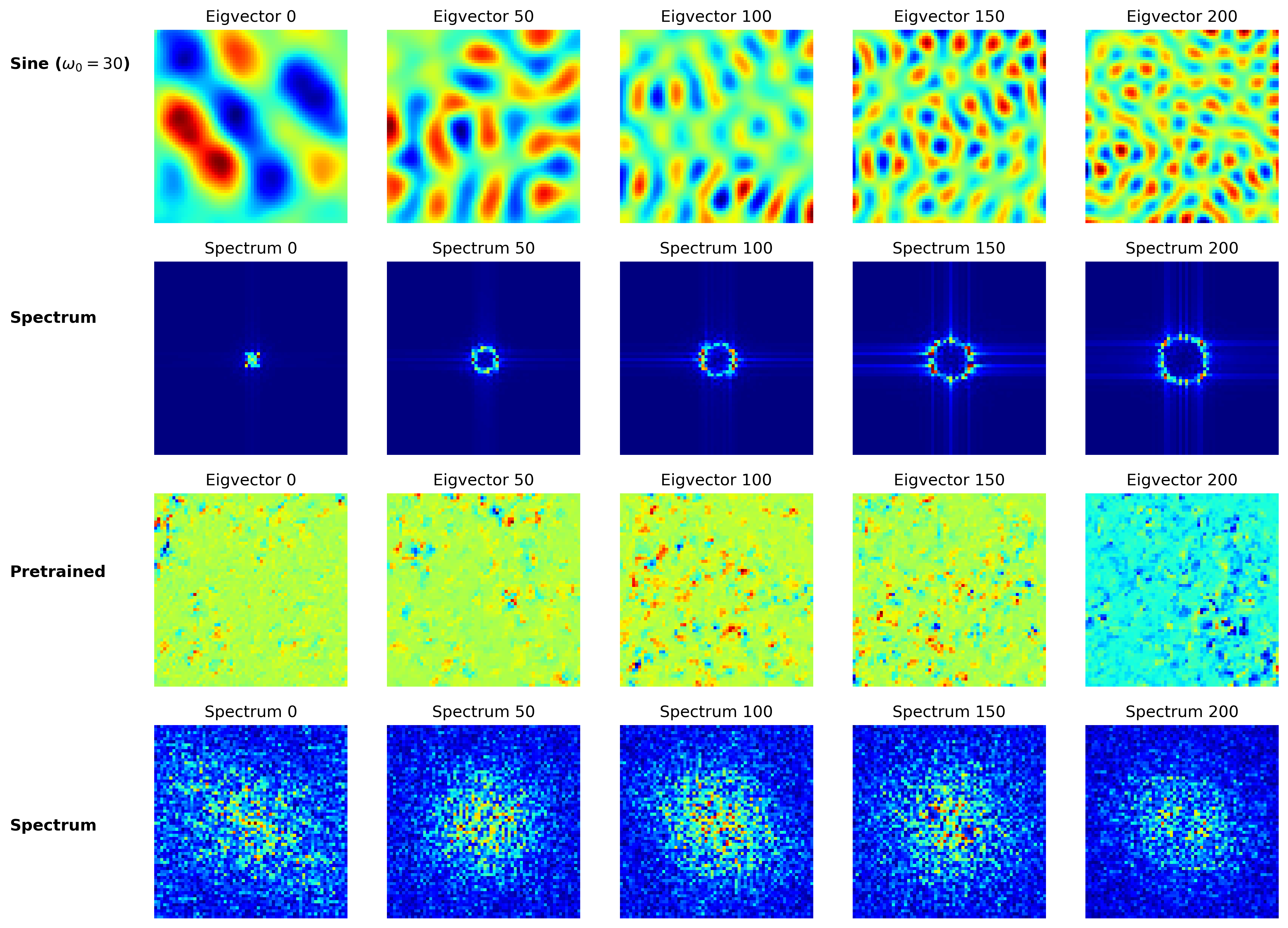}
    \caption{Comparison of the top eigenvectors of the NTK for a network initialized with standard random weights and a network initialized with noise as target self-supervised pretraining and the corresponding spectrum.}
    \label{fig: eigvecs}
 \end{figure*}

\section{Theory}\label{sec:Theory}
In this section, we introduce a noise-target self-supervised initialization framework and develop a unified theoretical framework to investigate its impact on optimization dynamics. 

\subsection{Problem setup}\label{sec:background}
Let $f(x; \theta): \mathbb{R}^d \rightarrow \mathbb{R}^m$ denote a neural network parameterized by $\theta \in \mathbb{R}^p$.
Given a dataset of input-target pairs $\{(x_i, y_i)\}_{i=1}^n$, the training objective is to minimize the empirical risk
\begin{equation}
L(\theta) = \frac{1}{n} \sum_{i=1}^n l(f(x_i; \theta), y_i),
\label{eq:loss_function}
\end{equation}
where $l$ denotes the loss function, and $\theta$ is typically optimized using gradient-based optimization methods.

The initial parameters $\theta_0$ play a critical role in shaping the descent trajectory of gradient methods.
This effect is especially strong in over-parameterized models, which often have highly non-convex loss landscapes with large regions where gradients are close to zero \cite{glorot2010understanding,He2015Delving,du2019gradient}.
Our objective is to design an initialization strategy for the construction of $\theta_0$ that improves the efficiency of optimization without relying on labeled data or semantic priors.

\subsection{Noise-driven self-supervised initialization}\label{sec:noise_pretrain}
Our method decouples the initialization procedure from any task-specific data by performing a preliminary pretraining phase on random noise.
Let $\varepsilon$ denote a random noise target image, we define a self-supervised objective function as
\begin{equation}
\theta_0
=
\arg\min_{\theta}
l\big(f(x;\theta), \varepsilon\big),
\qquad
\varepsilon \sim p(\varepsilon).
\label{eq:noise}
\end{equation}
where $p(\varepsilon)$ denotes a noise distribution, which may be
Gaussian or uniform depending on the range of values of the targets.
Compared with equation \eqref{eq:loss_function}, this objective simply replaces the original labels $y$ with random noise $\varepsilon$.

Optimizing Equation \eqref{eq:noise} induces a structured starting point in parameter space.
This noise pre-training initializes a parameter state for downstream tasks.
Unlike transfer learning or fine-tuning, our approach does not rely on semantically meaningful signals during the pretraining stage.
Moreover, it does not introduce architectural constraints and incurs minimal computational overhead.

\subsection{Neural Tangent Kernel (NTK) analysis}\label{sec:ntk}
In this section, we analyze the effect of noise-driven self-supervised pretraining on optimization dynamics through the lens of NTK theory \cite{jacot2018neural} on implicit neural representation (INR) models \cite{sitzmann2020implicit}.
While prior work on accelerating INR training has focused primarily 
 on modifying activation functions \cite{sitzmann2020implicit, liu2023finer, saragadam2023wire} or input encodings \cite{tancik2020fourier,Thomas2022Instant}, our approach instead targets the initialization of network weights.
To the best of our knowledge, the topic of a universal pretraining framework for initialization is still a largely open problem.

The NTK framework provides a principled tool for analyzing the training dynamics of INRs by expanding the network output $f(x; \theta)$ around the initialization $\theta_0$ via a first-order Taylor expansion:
\begin{equation}
f(x; \theta) \approx f(x; \theta_0) + \nabla_{\theta} f(x; \theta_0) (\theta - \theta_0).
\end{equation}
Under this approximation, the training dynamics of the network can be described by the NTK, defined as
\begin{equation}
K(x, x') = \nabla_{\theta} f(x; \theta_0) \nabla_{\theta} f(x'; \theta_0)^T.
\label{eq: ntk}
\end{equation}

Although NTK theory is exact only in the network layer infinite-width limit, it has been shown to provide meaningful qualitative insights into the training dynamics of sufficiently wide finite networks \cite{ortizjimenez2021what}.
By performing an eigen-decomposition of the NTK
\begin{equation}
K = U \Lambda U^{\mathrm{T}}, \quad
\Lambda = \mathrm{diag}(\lambda_1, \dots, \lambda_n), \quad
U = [u_1, \dots, u_n].
\end{equation}
where the eigenvectors $u_i$ correspond to the principal modes of network output, and the associated eigenvalues $\lambda_i$ govern 
the convergence rate along these directions. 
Larger eigenvalues correspond to faster learning (convergence) along that direction, while smaller eigenvalues correspond to slower learning.
This decomposition helps explain the network's implicit bias in fitting different components of the target function.

 \begin{figure}[htbp]
\centering
    \includegraphics[width=0.5\columnwidth]
    {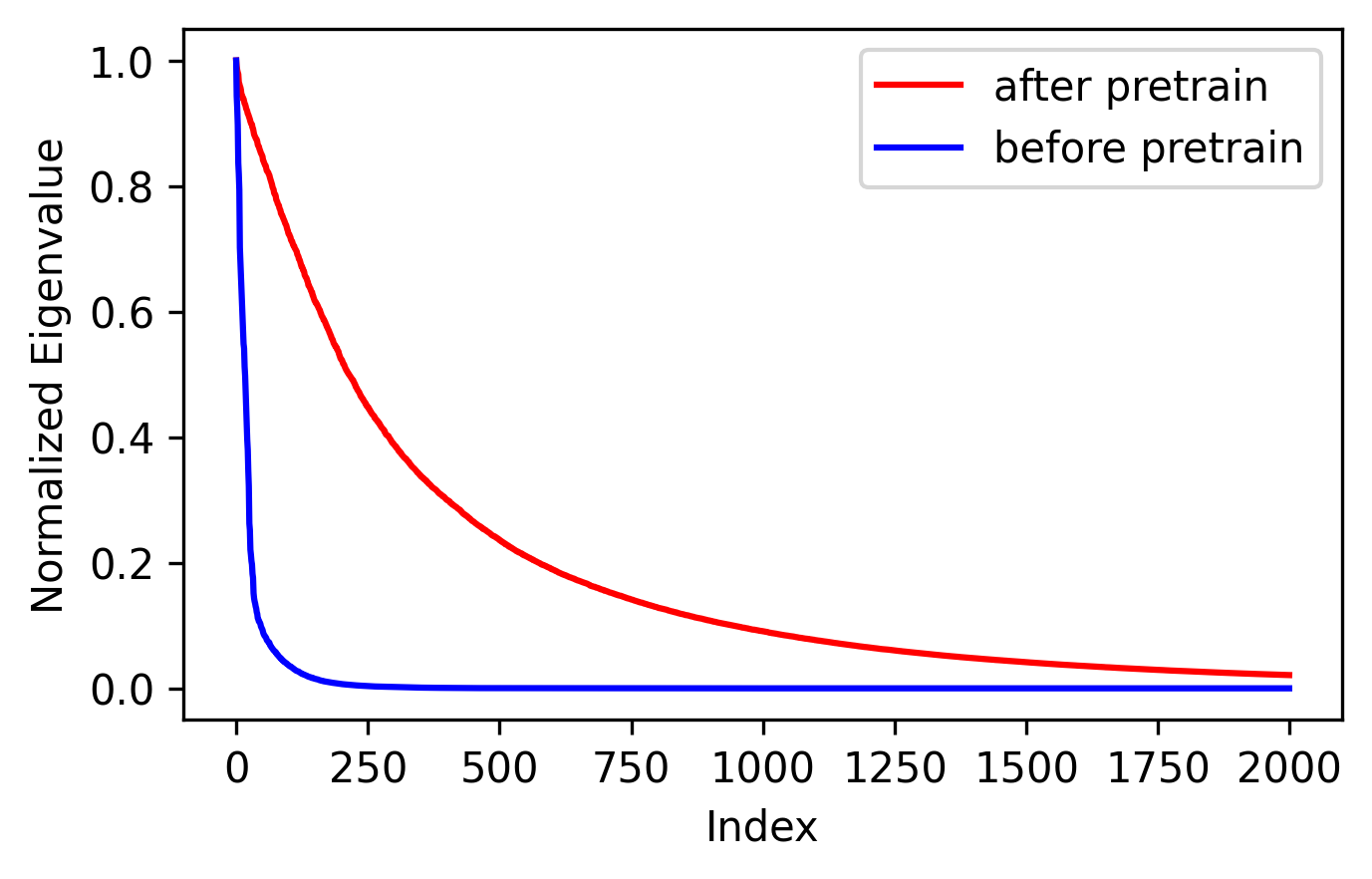}
    \caption{Comparison of the NTK eigenvalue for Siren having standard random weights initialization with noise-driven self-supervised pretraining.}
    \label{fig: eigvalues}
 \end{figure}

\begin{figure}[htbp]
\centering
    \includegraphics[width=\columnwidth]
    {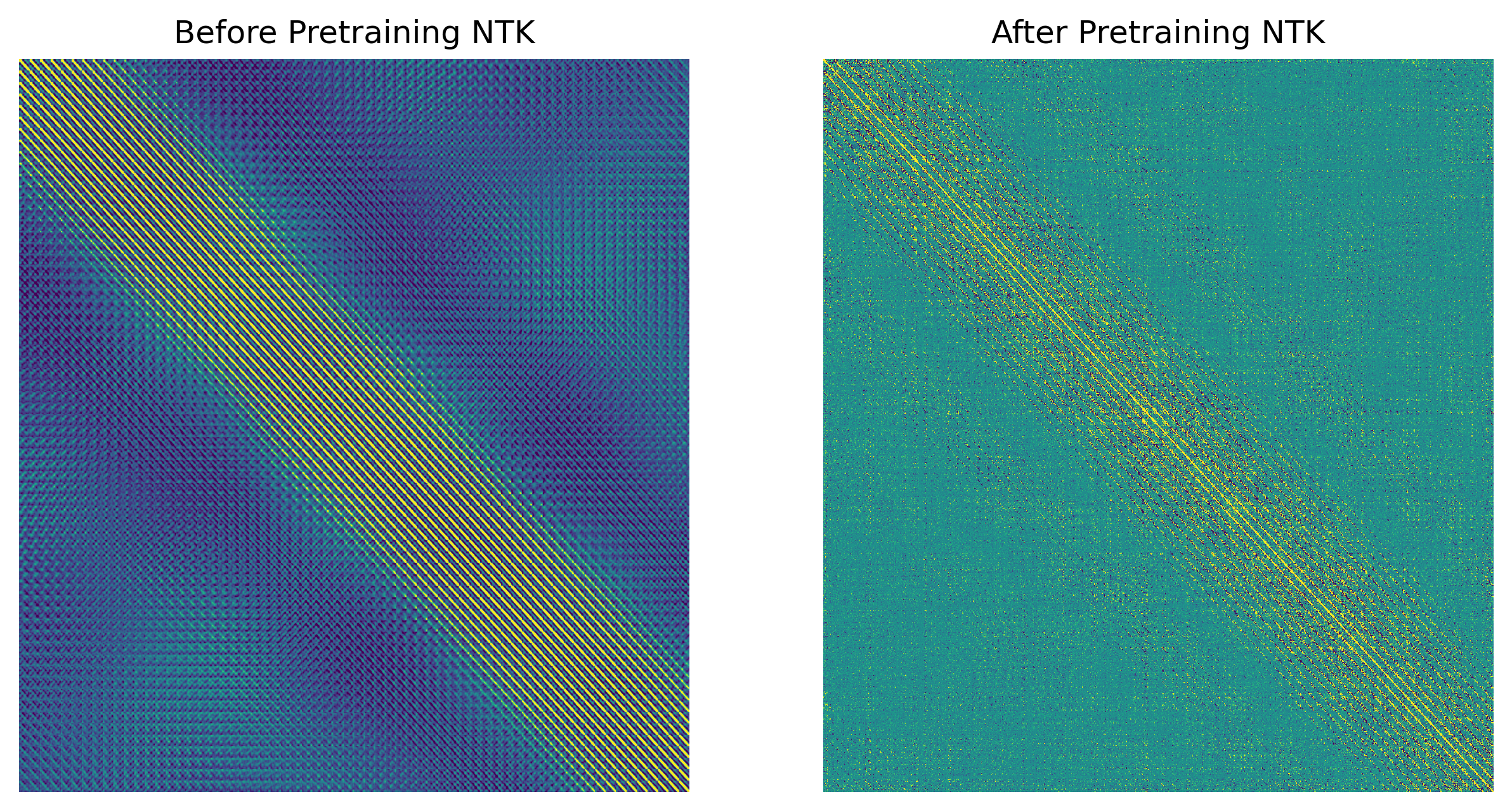}
    \caption{Comparison of the NTK matrix based on Siren with standard random weights and with noise-driven self-supervised pretraining.}
    \label{fig: ntk}
 \end{figure}
 
 \begin{figure*}[htbp]
\centering
    \includegraphics[width=0.9\textwidth]
    {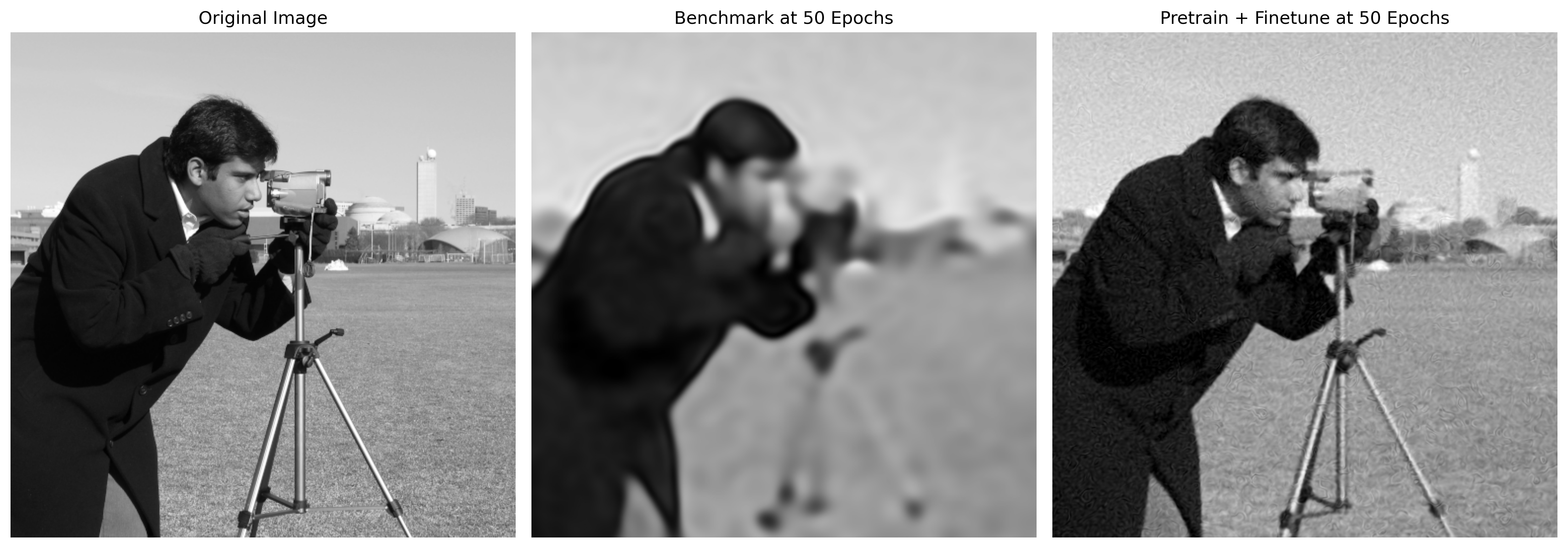}
    \caption{Ground truth image and comparison of the image representation results after 50 iterations of training with standard random initialization and with noise-driven self-supervised pretraining using in both cases the same network and training hyperparameters.}
    \label{fig: inr_50}
 \end{figure*}
 
 \begin{figure}[htbp]
\centering
    \includegraphics[width=\columnwidth]
    {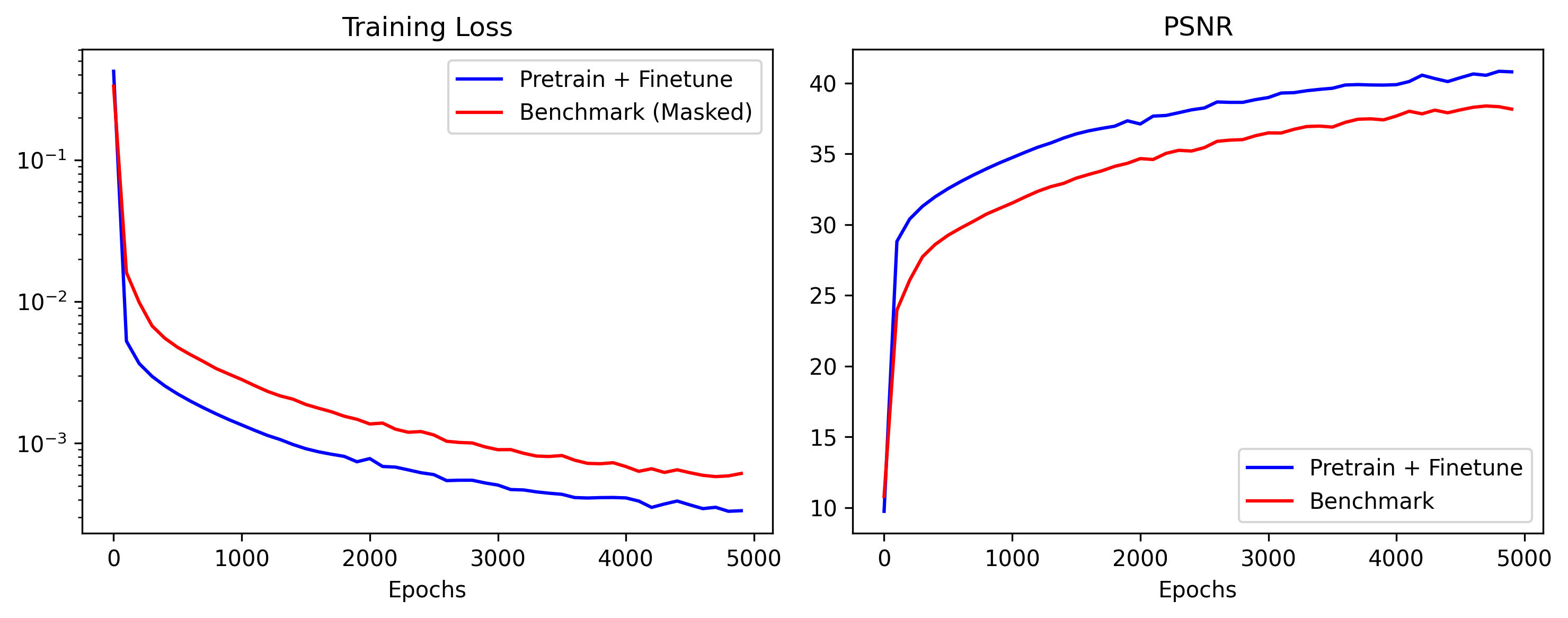}
    \caption{The loss and PSNR curves of the image representation task with standard random initialization and with noise-driven self-supervised pretraining.}
    \label{fig: inr_metrics}
 \end{figure}

Figure~\ref{fig: eigvecs} shows the NTK of a SIREN network \cite{sitzmann2020implicit} with 3 layers and 256 neurons per layer and its eigenspectrum before and after random noise pre-training. 
For the standard SIREN network, the eigenvectors in the frequency domain exhibit a gradually expanding ring structure (Figure~\ref{fig: eigvecs} second row).
This indicates that the network possesses an intrinsic bias towards representing signals with progressively higher frequencies, consistent with the harmonic structure induced by the sinusoidal activation.
The corresponding eigenvectors reflect this frequency hierarchy, with lower-frequency components dominating the leading eigenmodes and higher-frequency components appearing in successive modes. As analyzed in \citep{yuce2022structured}, the eigenvectors of the NTK can be regarded as a dictionary of basis functions that the network uses to represent input signals, so under the original initialization, the frequencies of the eigenvectors gradually increase, which provides evidence that SIREN gradually learns the signal  from low to high frequencies.

In contrast, noise pretraining deconstructs this intrinsic bias.
The resulting eigenvectors (Figure~\ref{fig: eigvecs}, third row) show a disordered noise-like eigen-spectrum in the frequency domain, with no preference for specific frequencies.
Accordingly, the eigenvectors of the pretrained model are devoid 
of any systematic spectral ordering, providing random projections.
This implies that the pretraining process effectively weakens the 
network initialization from its original sinusoidal prior, removing the structured frequency-specific learning dynamics and producing a kernel that is spectrally flatter and less biased.

As shown in Figure~\ref{fig: eigvalues}, before pretraining, the NTK eigenvalues of SIREN rapidly decay, indicating that the network naturally prioritizes the dominant directions, typically corresponding to low-frequency components.
This spectral bias induces a characteristic coarse-to-fine learning behavior, whereby the network initially fits the global, low-frequency structure of the signal before gradually incorporating higher-frequency details.
After pretraining on random noise, the eigenvalue decay becomes significantly slower, suggesting that the energy of the gradient is distributed more evenly across different modes of the NTK.
As a result, the pretrained network exhibits substantially weakened frequency ordering across NTK modes, reducing the dominance of low-frequency components during early optimization and attenuating the strong coarse-to-fine learning behavior observed in the standard SIREN.

As shown in Figure~\ref{fig: ntk}, the pretrained NTK matrix exhibits a narrower bright band along the diagonal than the original, indicating a stronger locality and a higher frequency selectivity of the kernel.
A narrower diagonal band implies that strong correlations are confined to nearby input locations, with correlations rapidly decaying as the input distance increases.
This implies that the network is more sensitive to fine- or high-frequency structures.

From an optimization perspective, such a localized NTK allows gradient updates to focus on nearby samples, which can accelerate the learning of detailed features and high-frequency components.
However, increased locality may also reduce robustness, as small perturbations in the data can have a stronger influence on the training dynamics.
In contrast, a wider bright band corresponds to weaker locality and a smoother kernel, where correlations extend over larger input distances.
This kernel promotes more stable optimization and favors learning of global or low-frequency structures, but may slow the convergence of high-frequency details, consistent with observations in \cite{sitzmann2020implicit}.

Overall, the width of the diagonal bright band reflects a trade-off between stability and expressiveness: narrower bands enhance sensitivity to fine-scale details at the cost of robustness, whereas wider bands promote smoother and more stable optimization but limit the efficient fitting of fine-scale details.

\begin{figure*}[htbp]
\centering
    \includegraphics[width=0.9\textwidth]
    {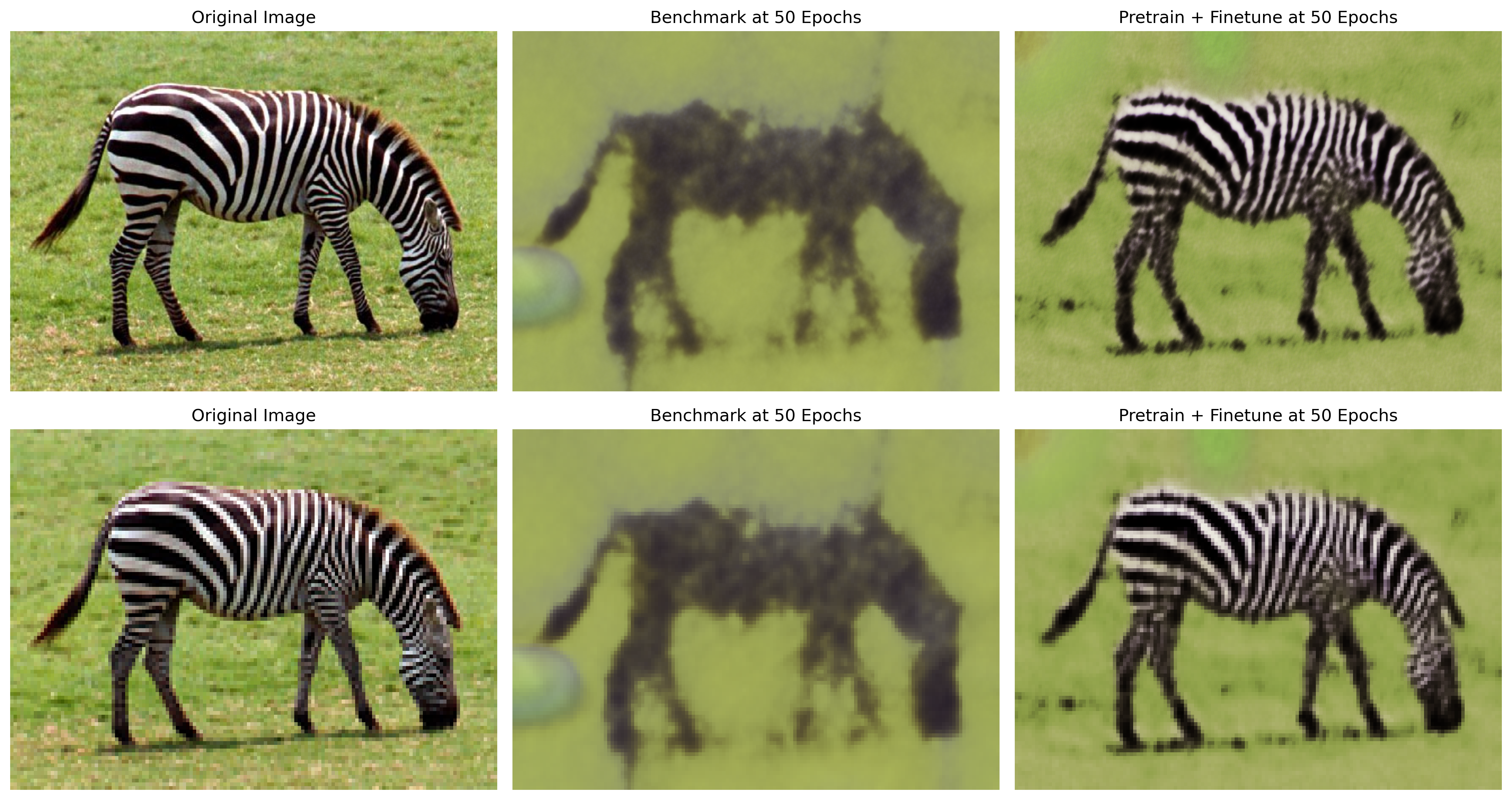}
    \caption{Comparison of the super-resolution results after $50$ iterations of DIP training with standard random initialization and with noise-driven self-supervised pretraining.
    The first row shows high-resolution ground truth images, random initialization after $50$ iterations and pretrained initialization after 50 iterations.
    The second row shows the corresponding low-resolution input images.}
    \label{fig: dip_sr_first50}
 \end{figure*}
 
 \begin{figure*}[htbp]
\centering
    \includegraphics[width=0.9\textwidth]
    {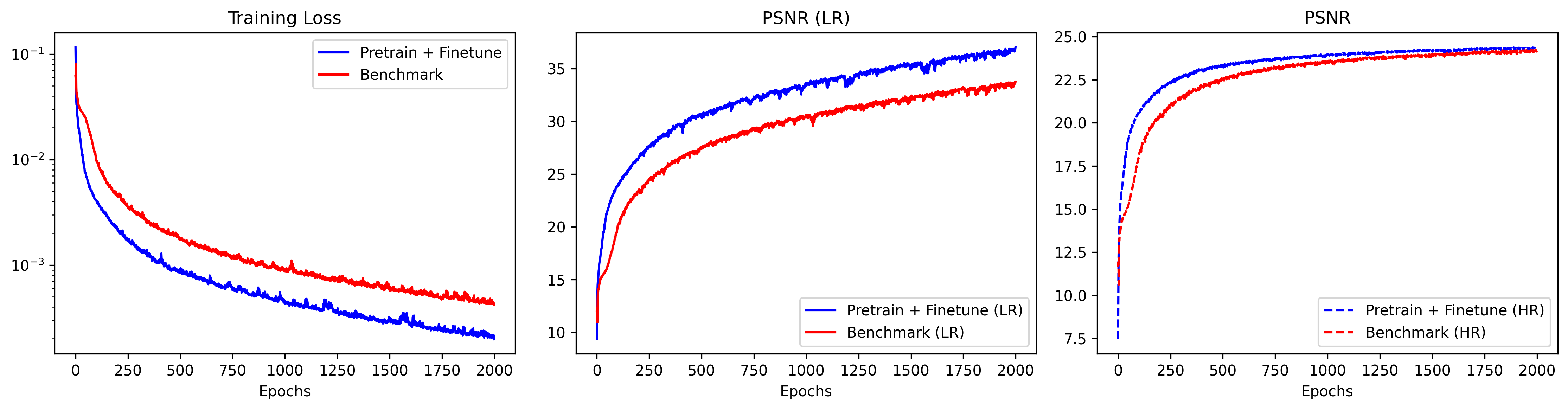}
    \caption{Comparison of the loss curves and PSNR curves on the super-resolution task with standard random initialization and with noise-driven self-supervised pretraining.}
    \label{fig: dip_sr_metrics}
 \end{figure*}

\section{Experiments}\label{sec:Experiments}
In this section, we empirically evaluate the proposed noise-driven self-supervised initialization strategy on INR and deep image prior (DIP) tasks, including image representation, image denoising, inpainting and super-resolution.
These experiments are designed to assess both the effectiveness and general applicability of the proposed initialization method when applied to convolutional neural network (CNN) architectures.

\subsection{Coordinate-based image representation}\label{sec:image_representation}
The coordinate-based image representation task aims to learn a continuous mapping from 2D spatial coordinates to RGB pixel values based on the following loss function
\begin{equation}
L(\theta) = \| f(x; \theta) - y \|_2^2,
\end{equation}
where $x$ is the 2D coordinate, $y$ is the corresponding RGB pixel value, and $f(x; \theta)$ is a sine activated MLP parameterized by $\theta$ that maps spatial coordinates to pixel values.
A grayscale image of size $512 \times 512$ is used as the target image for this task.
During training, all pixel coordinates are utilized as training samples in each optimization iteration.
For the noise-driven self-supervised pretraining phase, Gaussian noise with zero mean and a standard deviation of $1.0$ is employed as the regression target.
The initialization of the baseline weight for the MLP follows the scheme proposed in the original SIREN paper \cite{sitzmann2020implicit}.
For the proposed pretrained initialization, the network starts with the same baseline weights and is first pretrained on random noise targets for $200$ iterations using the Adam optimizer with a learning rate of $1\times10^{-4}$.

 \begin{figure*}[htbp]
\centering
    \includegraphics[width=0.9\textwidth]
    {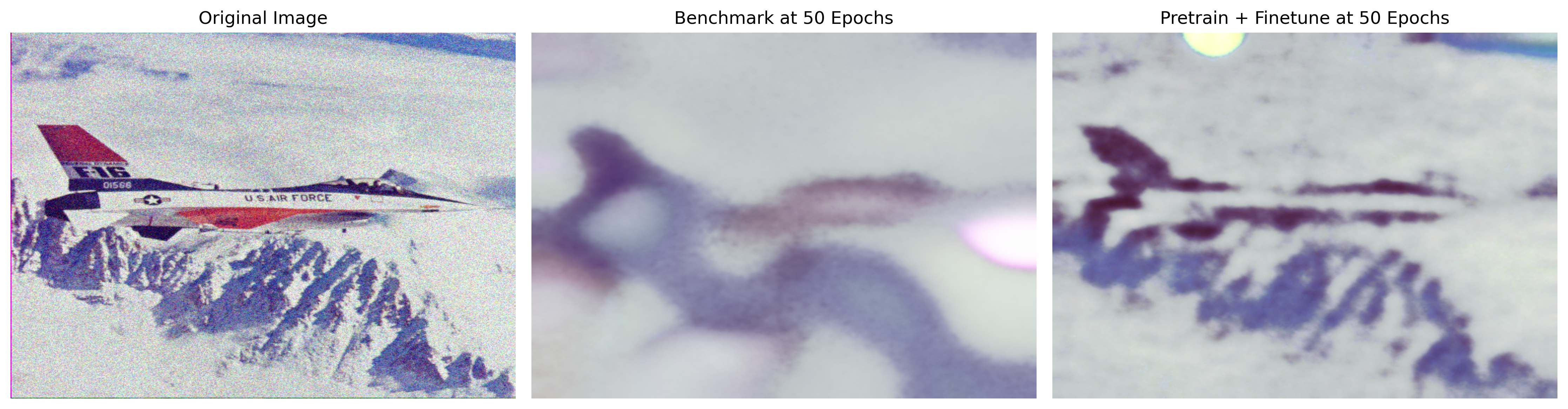}
    \caption{The noisy image, denoised results after 50 iterations of DIP training with standard random initialization and with noise-driven self-supervised pretraining.}
    \label{fig: dip_denoising_first50}
 \end{figure*}
 
 \begin{figure*}[htbp]
\centering
    \includegraphics[width=0.9\textwidth]
    {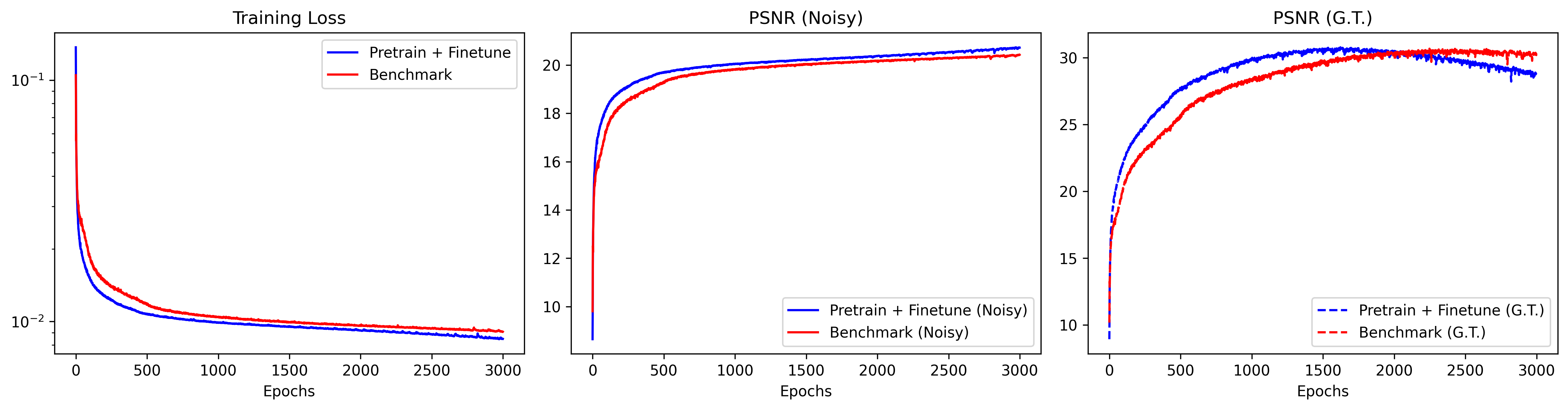}
    \caption{Comparison of the loss curves and PSNR curves on the denoising task with standard random initialization and with noise-driven self-supervised pretraining.}
    \label{fig: dip_denoising_metrics}
 \end{figure*}

Figure~\ref{fig: inr_50} compares the reconstruction results obtained after $50$ training iterations using the benchmark random initialization and the proposed noise-driven self-supervised pretraining initialization.
With the standard initialization, the network struggles to recover high-wavenumber components at this early stage, resulting in overly smooth reconstructions.
This behavior is consistent with the well-known spectral bias of SIREN networks, which preferentially learn low-frequency components during the initial phase of optimization.
In contrast, the proposed initialization strategy substantially enhances the recovery of high-wavenumber components at the early stages of the optimization.

Figure~\ref{fig: inr_metrics} presents the training loss and PSNR curves for the image representation task using standard random initialization and the proposed noise-driven self-supervised pretraining.
Although the pretrained model starts with a higher loss and a lower PSNR due to the random noise pretraining targets, it rapidly surpasses the benchmark initialization within the first $100$ training iterations.
Moreover, the pretrained initialization consistently outperforms the benchmark initialization throughout the entire training process, demonstrating its effectiveness in accelerating convergence and improving early-stage reconstruction quality.

\subsection{Super-resolution}\label{sec:super-resolution}
The super-resolution task aims to reconstruct a high-resolution (HR) image from a low-resolution (LR) observation based on the following loss function
\begin{equation}
L(\theta) = \| D(f(x; \theta)) - y \|_2^2,
\end{equation}
where $D$ denotes a down-sampling operator, $y$ is the LR image, and $f(x; \theta)$ is a CNN that maps a fixed random input $x$ to the reconstructed HR image.
Before training on the super-resolution task, the CNN is pretrained using random noise targets, 
Specifically, uniform noise sampled from the range $[0, 1]$ is used to match the dynamic range of the image pixel values.

The network architecture and hyperparameter settings for the super-resolution task follow those described in \cite{ulyanov2018deep}.
During the pretraining stage, the Adam optimizer is employed with a learning rate of $0.01$ for $500$ iterations.
Figure~\ref{fig: dip_sr_first50} compares the reconstruction results after $50$ training iterations using the original DIP framework with standard random initialization and the proposed noise-driven self-supervised initialization.
The proposed initialization strategy significantly improves the convergence speed of the network, with high-frequency details being more effectively recovered during the early stages of optimization.

Figure~\ref{fig: dip_sr_metrics} shows the training loss and PSNR curves for the super-resolution task under standard random initialization and noise-driven self-supervised pretraining.
As evidenced by the curves, the proposed initialization strategy substantially accelerates network convergence, particularly during the early optimization phase.
Moreover, the PSNR curves indicate that the proposed method consistently achieves higher reconstruction quality at the same number of training iterations compared to standard random initialization.

\subsection{Denosing}\label{sec:denosing}
The denoising task aims to recover a clean image from a noisy observation based on the following loss function
\begin{equation}
L(\theta) = \| f(x; \theta) - y \|_2^2,
\end{equation}
where $y$ is the noisy image, and $f(x; \theta)$ is a CNN that maps a fixed random input $x$ to an image estimate, which is expected to approximate the underlying clean image.

The network architecture and hyperparameter settings for the denoising task follow those described in \cite{ulyanov2018deep}.
During the pretraining stage, the Adam optimizer is employed with a learning rate of $1\times10^{-2}$ for $1000$ iterations.
Figure~\ref{fig: dip_denoising_first50} compares the reconstruction results after $50$ training iterations using the original DIP framework with standard random initialization and the proposed noise-driven self-supervised initialization.
Similarly to the super-resolution task, under standard random initialization, the network primarily reconstructs coarse background structures during the early training stage.
In contrast, the proposed initialization strategy enables the network to recover finer image details at an earlier stage of the optimization.

Figure~\ref{fig: dip_denoising_metrics} presents the training loss and PSNR curves for the denoising task under standard random initialization and noise-driven self-supervised pretraining.
Similar to the super-resolution task, the proposed initialization strategy significantly accelerates the convergence of the network.
However, when evaluated against the ground-truth image, the PSNR exhibits a different trend, beginning to decline after approximately $1500$ training iterations.
Noise-driven pretraining enables the network to reach its peak PSNR at an earlier stage of training.
Beyond this point, the PSNR gradually decreases as the network starts to overfit the noise present in the observation.
In contrast, the conventional initialization has not yet reached this overfitting regime within the same number of iterations and continues to exhibit gradual improvement.
This observation highlights the importance of early stopping when employing DIP-based denoising methods, particularly when using our accelerated initialization strategies.

\subsection{Inpainting}\label{sec:inpainting}
The inpainting task aims to reconstruct missing regions of an image from partial observations based on the following loss function
\begin{equation}
L(\theta) = \| M \odot (f(x; \theta) - y) \|_2^2,
\end{equation}
where $M$ is a binary mask indicating the observed pixel locations, $y$ denotes the corrupted image with missing regions, and $f(x; \theta)$ is a CNN that maps a fixed random input $x$ to the complete image estimate.

The network architecture and hyperparameter settings for the inpainting task follow those described in \cite{ulyanov2018deep}.
During the pretraining stage, the Adam optimizer is employed with a learning rate of $1\times10^{-2}$ for $500$ iterations. 
 \begin{figure*}[htbp]
\centering
    \includegraphics[width=0.9\textwidth]
    {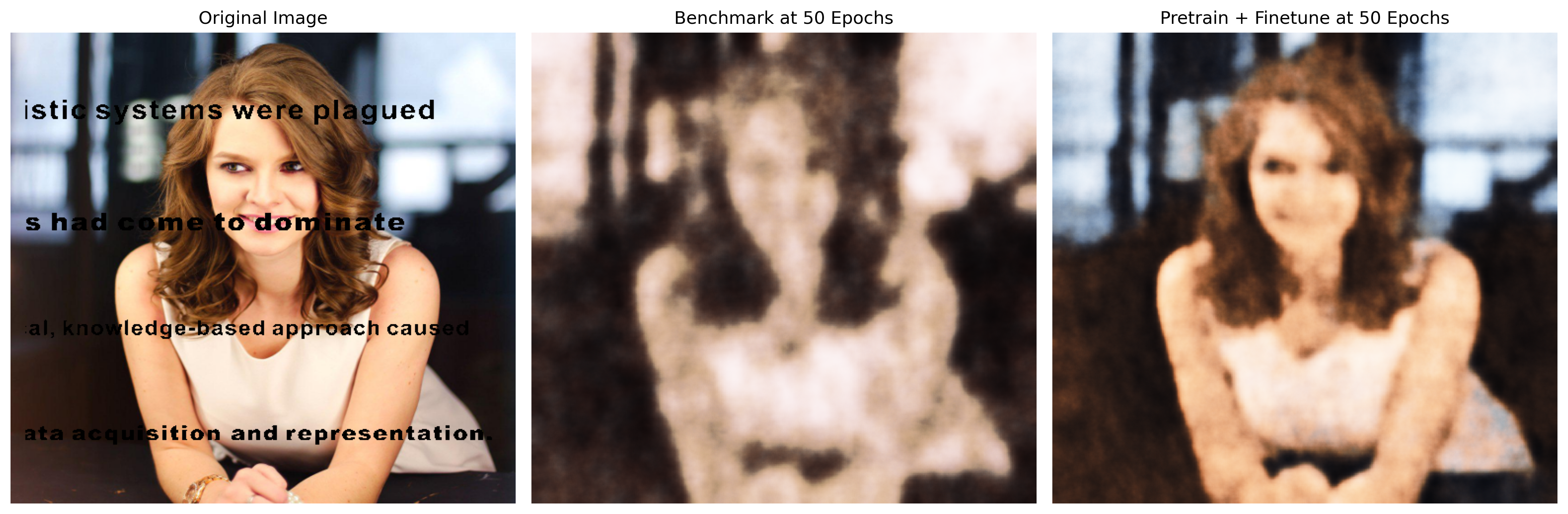}
    \caption{The corrupted image with blurred regions, inpainted results after 50 iterations of DIP training with a standard random initialization and with noise-driven self-upervised pretraining.}
    \label{fig: dip_inpainting_first50}
 \end{figure*}

 \begin{figure*}[htbp]
\centering
    \includegraphics[width=0.9\textwidth]
    {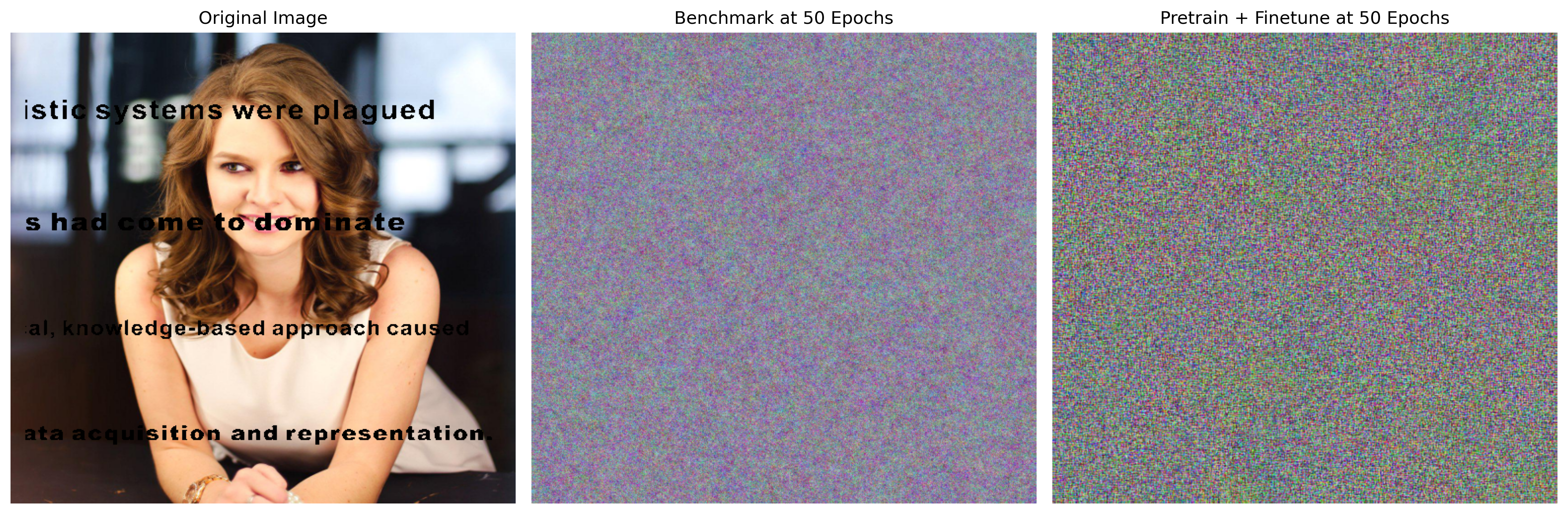}
    \caption{The corrupted image with blurred regions, the initial output of the neural network of a standard random initialization and the pretrained initialization.}
    \label{fig: dip_inpainting_first0}
 \end{figure*}

  \begin{figure*}[htbp]
\centering
    \includegraphics[width=0.9\textwidth]
    {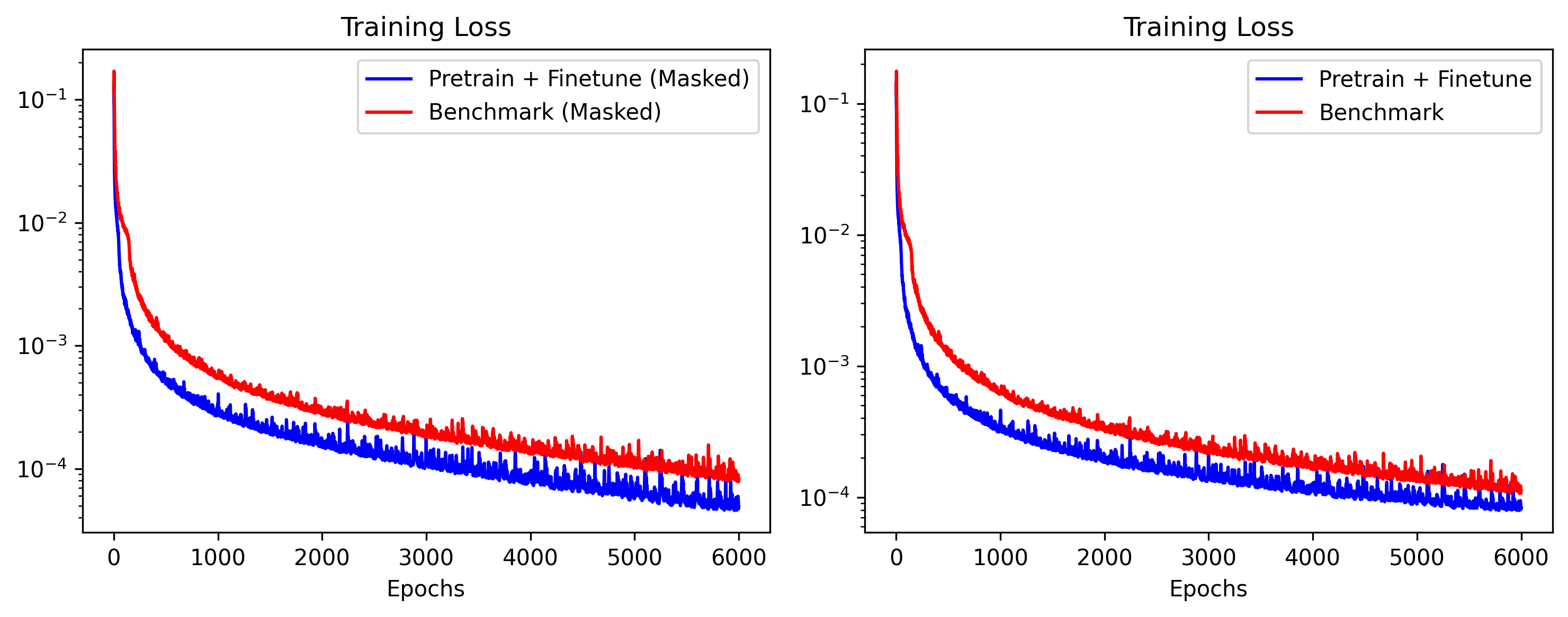}
    \caption{The corrupted image with blurred regions, the initial output of the neural network of standard random initialization and pretrained initialization.}
    \label{fig: dip_inpainting_metrics}
 \end{figure*}

In this example, the wavenumber content of the results after 50 iterations (\ref{fig: dip_inpainting_first50}) appears very similar, only the brightness is slightly different. This is because, when we examine the outputs at the beginning (\ref{fig: dip_inpainting_first0}), the original initialization and our proposed initialization, we find that both already contain substantial high-frequency components. As a result, the original initialization already gives the neural network some ability to learn high-frequency information. Nonetheless, the metrics shown in \ref{fig: dip_inpainting_metrics} still indicate that our method can accelerate convergence.

\section{Conclusion}\label{sec:Conclusion}
We introduced noise-based target self-supervised pretraining as a structured initialization method for neural networks. Our empirical analysis demonstrates that random noise, despite containing no semantic information, provides a useful self-supervised signal that guides the network towards structured weight configurations. This initialization accelerates convergence, stabilizes early-stage optimization, and reduces sensitivity to hyperparameters.
From a theoretical perspective, we analyzed the method within the Neural Tangent Kernel (NTK) framework and showed that noise-based pretraining shapes the NTK at initialization, improving spectral properties that control convergence rates without changing the overall training dynamics. Although NTK assumes infinitely wide networks, and thus does not fully account for feature learning in finite networks, empirical results indicate that the benefits persist in practical scenarios.
Future work could explore the effectiveness of noise-driven pretraining in diffusion models, large-scale supervised learning tasks and also the potential in accelerating the convergence of physics-informed neural networks (PINNs), investigate more structured noise distributions and self-supervised objectives, and further study how this approach can improve optimization and convergence in these broader settings.

\bibliographystyle{unsrt}  

\bibliography{references}

@inproceedings{He2015Delving,
  title		=	{Delving Deep into Rectifiers: Surpassing Human-Level Performance on ImageNet Classification},
  author	=	{Kai-Ming He and Xiang-Yu Zhang and Shao-Qing Ren and Jian Sun},
  booktitle	=	{IEEE International Conference on Computer Vision},
  volume	=	{},
  number	=	{},
  pages		=	{1026--1034},
  year		=	{2015},
}

@inproceedings{glorot2010understanding,
  title		=	{Understanding the difficulty of training deep feedforward neural networks},
  author	=	{Xavier Glorot and Yoshua Bengio},
  booktitle	=	{Proceedings of the thirteenth international conference on artificial intelligence and statistics},
  volume	=	{},
  number	=	{},
  pages		=	{249--256},
  year		=	{2010},
  doi		= 	{}
}

@inproceedings{doersch2015unsupervised,
  title		=	{Unsupervised visual representation learning by context prediction},
  author	=	{Carl Doersch and Abhinav Gupta and Alexei A Efros},
  booktitle	=	{IEEE International Conference on Computer Vision},
  volume	=	{},
  number	=	{},
  pages		=	{1422--1430},
  year		=	{2015},
  doi		= 	{}
}

@inproceedings{chen2020simple,
  title		=	{A simple framework for contrastive learning of visual representations},
  author	=	{Ting Chen and Simon Kornblith and Mohammad Norouzi and Geoffrey Hinton},
  booktitle	=	{International conference on machine learning},
  volume	=	{},
  number	=	{},
  pages		=	{1597--1607},
  year		=	{2020},
  doi		= 	{}
}

@article{jacot2018neural,
  title		=	{Neural tangent kernel: Convergence and generalization in neural networks},
  author	=	{Arthur Jacot and Franck Gabriel and Cl{\'e}ment Hongler},
  journal	=	{Advances in neural information processing systems},
  volume	=	{31},
  number	=	{},
  pages		=	{},
  year		=	{2018},
}

@inproceedings{du2019gradient,
  title		=	{Gradient descent provably optimizes over-parameterized neural networks},
  author	=	{Simon S Du and Xi-Yu Zhai and Barnabas Poczos and Aarti Singh},
  booktitle	=	{International Conference on Learning Representations},
  volume	=	{},
  number	=	{},
  pages		=	{1026--1034},
  year		=	{2019},
}

@inproceedings{ulyanov2018deep,
  title		=	{Deep image prior},
  author	=	{Dmitry Ulyanov and Andrea Vedaldi and Victor Lempitsky},
  booktitle	=	{IEEE conference on computer vision and pattern recognition},
  volume	=	{},
  number	=	{},
  pages		=	{9446--9454},
  year		=	{2018},
}

@article{sitzmann2020implicit,
  title		=	{Implicit neural representations with periodic activation functions},
  author	=	{Vincent Sitzmann and Julien Martel and Alexander Bergman and David Lindell and Gordon Wetzstein},
  journal	=	{Advances in neural information processing systems},
  volume	=	{33},
  number	=	{},
  pages		=	{7462--7473},
  year		=	{2020},
}

@article{tancik2020fourier,
  title		=	{Fourier features let networks learn high frequency functions in low dimensional domains},
  author	=	{Matthew Tancik and Pratul Srinivasan and Ben Mildenhall and Sara Fridovich-Keil and Nithin Raghavan and Utkarsh Singhal and Ravi Ramamoorthi and Jonathan Barron and Ren Ng},
  journal	=	{Advances in neural information processing systems},
  volume	=	{33},
  number	=	{},
  pages		=	{7537--7547},
  year		=	{2020},
}

@article{Thomas2022Instant,
  title		=	{Instant neural graphics primitives with a multiresolution hash encoding},
  author	=	{Thomas M{\"u}ller and Alex Evans and Christoph Schied and Alexander Keller},
  journal	=	{ACM transactions on graphics (TOG)},
  volume	=	{41},
  number	=	{4},
  pages		=	{1--15},
  year		=	{2022},
}

@article{liu2023finer,
  title   = {FINER: Flexible Spectral-Bias Tuning in Implicit Neural Representation by Variable-periodic Activation Functions},
  author  = {Zhen Liu and Hao Zhu and Qi Zhang and Jingde Fu and Weibing Deng and Zhan Ma and Yanwen Guo and Xun Cao},
  journal = {arXiv},
  volume  = {},
  number  = {},
  pages   = {},
  year    = {2023},
}

@article{saragadam2023wire,
  title   = {WIRE: Wavelet Implicit Neural Representations},
  author  = {Vishwanath Saragadam and Daniel LeJeune and Jasper Tan and Guha Balakrishnan and Ashok Veeraraghavan and Richard G. Baraniuk},
  journal = {arXiv},
  volume  = {},
  number  = {},
  pages   = {},
  year    = {2023},
}

@article{yuce2022structured,
  title   = {A Structured Dictionary Perspective on Implicit Neural Representations},
  author  = {Gizem Y{\"u}ce and Guillermo Ortiz-Jim{\'e}nez and Beril Besbinar and Pascal Frossard},
  journal = {arXiv},
  volume  = {},
  number  = {},
  pages   = {},
  year    = {2022},
}

@article{cheon2025pretraining,
  title   = {Pretraining with Random Noise for Fast and Robust Learning without Weight Transport},
  author  = {Cheon, Jeonghwan and Lee, Sang Wan and Paik, Se-Bum},
  journal = {arXiv preprint arXiv:2405.16731},
  year    = {2025},
  volume  = {},
  number  = {},
  pages   = {},
  doi     = {10.48550/arXiv.2405.16731}
}

@article{lillicrap2016random,
  title   = {Random synaptic feedback weights support error backpropagation for deep learning},
  author  = {Lillicrap, Timothy P. and Cownden, Daniel and Tweed, Douglas B. and Akerman, Colin J.},
  journal = {Nature Communications},
  volume  = {7},
  pages   = {13276},
  year    = {2016},
  doi     = {10.1038/ncomms13276}
}

@article{ortizjimenez2021what,
  title   = {What can linearized neural networks actually say about generalization?},
  author  = {Ortiz-Jim{\'e}nez, Guillermo and Moosavi-Dezfooli, Seyed-Mohsen and Frossard, Pascal},
  journal = {arXiv preprint},
  volume  = {arXiv:2106.06770},
  year    = {2021},
  doi     = {10.48550/arXiv.2106.06770},
  url     = {https://arxiv.org/abs/2106.06770}
}





\end{document}